\documentclass[journal]{IEEEtran}

\usepackage{color}

\usepackage{hyperref}

\ifCLASSINFOpdf
  \usepackage[pdftex]{graphicx}
\else
  \usepackage[dvips]{graphicx}
  \graphicspath{{/figures}}
\fi
\usepackage{subcaption}

\usepackage{amsmath}
\usepackage{amssymb}
\usepackage{algorithm}
\usepackage{algorithmic}

\usepackage[inline]{enumitem}

\begin{document}

\title{An Incremental Self-Organizing Architecture \\ for Sensorimotor Learning and Prediction}

\author{Luiza Mici,~German I. Parisi,
        and~Stefan Wermter
\thanks{Authors are with the Department
of Informatics, Knowledge Technology, University of Hamburg, Vogt-Koelln-Strasse 30, 22527 Hamburg, Germany e-mail: {\tt\small \{mici, parisi, wermter\} @informatik.uni-hamburg.de}}
\thanks{Preprint submitted to IEEE Transactions on Cognitive and Developmental Systems (TCDS)}
}

\maketitle

\begin{abstract}
During visuomotor tasks, robots must compensate for temporal delays inherent in their sensorimotor processing systems. 
Delay compensation becomes crucial in a dynamic environment where the visual input is constantly changing, e.g., during the interacting with a human demonstrator.  
For this purpose, the robot must be equipped with a prediction mechanism for using the acquired perceptual experience to estimate possible future motor commands.
In this paper, we present a novel neural network architecture that learns prototypical visuomotor representations and provides reliable predictions on the basis of the visual input.
These predictions are used to compensate for the delayed motor behavior in an online manner.
We investigate the performance of our method with a set of experiments comprising a humanoid robot that has to learn and generate visually perceived arm motion trajectories. 
We evaluate the accuracy in terms of mean prediction error and analyze the response of the network to novel movement demonstrations.
Additionally, we report experiments with incomplete data sequences, showing the robustness of the proposed architecture in the case of a noisy and faulty visual sensor. 
\end{abstract}


\IEEEpeerreviewmaketitle

\section{Introduction}
\IEEEPARstart{R}{eal-time} interaction with the environment requires robots to adapt their motor behavior according to perceived events. 
However, each sensorimotor cycle of the robot is affected by an inherent latency introduced by the processing time of sensors, transmission time of signals, and mechanical constraints~\cite{mainprice2012sharing}\cite{Zhong2012}\cite{saegusa2007sensory}. 
Due to this latency, robots exhibit a discontinuous motor behavior which may compromise the accuracy and execution time of the assigned task. 

For social robots, delayed motor behavior makes human-robot interaction~(HRI) asynchronous and less natural.
Synchronization of movements during HRI may increase rapport and endow humanoid robots with the ability to collaborate with humans during daily tasks~\cite{lorenz2011synchronization}.
A possible solution to the sensorimotor latency is the application of predictive mechanisms which accumulate information from robot's perceptual and motor experience and learn an internal model which estimates possible future motor states~\cite{bahill1983simple}\cite{behnke2003predicting}.
The learning of these models in an unsupervised manner and their adaptation throughout the acquisition of new sensorimotor information remains an open challenge.

Latencies between perception and possible motor behavior occur in human beings~\cite{Nijhawan1063} as well. 
Such discrepancies are caused by neural transmission delays and are constantly compensated by predictive mechanisms in our sensorimotor system that account for both motor prediction and anticipation of the target movement. 
Miall et al.~\cite{miall1993cerebellum} have proposed that the human cerebellum is capable of estimating the effects of a motor command through an internal action simulation and a prediction model. 
Furthermore, there are additional mechanisms for visual motion extrapolation which account for the anticipation of the future position and movement of the target~\cite{kerzel2003neuronal}. 
Not only do we predict sensorimotor events in our everyday tasks, but we also constantly adjust our delay compensation mechanisms to the sensory feedback~\cite{rohde2014predictability} and to the specific task~\cite{deLaMalla}. 

Recently, there has been a considerable growth of learning-based prediction techniques, which mainly operate in a ``learn then predict" approach, i.e., typical motion patterns are extracted and learned from training data sequences and then learned patterns are used for prediction~\cite{Zhong2012}\cite{mainprice}\cite{ito2004line}\cite{levine2016learning}.
The main issue with this approach is that the adaptation of the learned models is interrupted by the prediction stage.
However, it is desirable for a robot operating in natural environments to be able to learn incrementally, i.e., over a lifetime of observations, and to refine the accumulated knowledge over time.
Therefore, the development of learning-based predictive methods accounting for both incremental learning and predictive behavior still need to be fully investigated. 

In this work, we propose a novel architecture that learns sensorimotor patterns and predicts the future motor states in an on-line manner. 
We evaluate the architecture in the context of an imitation task in an HRI scenario. 
In this scenario, body motion patterns demonstrated by a human demonstrator are mapped to trajectories of robot joint angles and then learned for subsequent imitation by a humanoid robot.
We approach the demonstration of the movements through motion capture with a depth sensor, which provides us with reliable estimations and tracking of a 3D human body pose.
Thus, the three-dimensional joint positions of the skeleton model constitute the input to the architecture.
The learning module captures spatiotemporal dependencies through a hierarchy of Growing When Required~(GWR)~\cite{marsland2002self}~networks, which has been successfully applied for the classification of human activities \cite{Parisi2016}\cite{Mici2016}.
The learning algorithm processes incoming robot joint angles and progressively builds a dictionary of motion segments.
Finally, an extended GWR algorithm, implemented at the last layer of our architecture, approximates a prediction function and utilizes the learned motion segments for predicting forthcoming motor commands.

We evaluate our system on a dataset of three subjects performing 10 arm movement patterns.
We study the prediction accuracy of our architecture while being continuously trained. 
Experimental results show that the proposed architecture can adapt quickly to an unseen pattern and can provide accurate predictions albeit continuously incorporating new knowledge.
Moreover, we show that the system can maintain its performance even when training takes place with missing sensory information. 

\section{Related Work}
\subsection{Motion prediction} \label{subseq:motion_prediction}
Motion analysis and prediction are an integral part of robotic platforms that counterbalance the imminent sensorimotor latency.
Well-known methods for tracking and prediction are the Kalman Filter models, as well as their extended versions which assume non-linearity of the system, and the Hidden Markov Models (HMM).
Kalman filter-based prediction techniques require a precise kinematic or dynamic model that describes how the state of an object evolves while being subject to a set of given control commands. HMMs describe the temporal evolution of a process through a finite set of states and transition probabilities.
Predictive approaches based on dynamic properties of the objects are not able to provide correct long-term predictions of human motion~\cite{vasquez2008intentional} due to the fact that human motion also depends on other higher-level factors than kinematic constraints, such as plans or intentions. 

There are some alternatives to approaches based on probabilistic frameworks in the literature and neural networks are probably the most popular ones. 
Neural networks are known to be able to learn universal function approximations and thereby predict non-linear data even though dynamic properties of a system or state transition probabilities are not known~\cite{schaefer2008learning}\cite{saegusa2007sensory}. 
For instance, Multilayer Perceptrons~(MLP)~and Radial Basis Function~(RBF)~networks as well as Recurrent Neural Networks have found successful applications as predictive approaches \cite{mainprice}\cite{barreto2007time}\cite{ito2004line}\cite{Zhong2012}. 
A subclass of neural network models, namely the Self-Organizing Map~(SOM)~\cite{kohonen1993self},~is able to perform local function approximation by partitioning the input space and learning the dynamics of the underlying process in a localized region. 
The advantage of the SOM-based methods is their ability to achieve long-term predictions at much less expensive computational time \cite{simon2007forecasting}. 

Johnson and Hogg~\cite{johnson1996learning} first proposed the use of multilayer self-organizing networks for the motion prediction of a tracked object. Their model consisted of a bottom SOM layer learning to represent the object states and the higher SOM layer learning motion trajectories through the leaky integration of neuron activations over time.  
Similar approaches were proposed later by Sumpter and Bulpitt~\cite{sumpter2000learning} and Hue et al.~\cite{hu2004learning}, who modeled time explicitly by adding lateral connections between neurons in the state layer, obtaining performances comparable to that of the probabilistic models. 

Several other approaches use SOMs extended with temporal associative memory techniques~\cite{barreto2007time}, e.g., associating to each neuron a linear Autoregressive (AR)~model~\cite{walter1990nonlinear}\cite{vesanto1997using}.
A drawback which is common to these approaches is their assumption of knowing a priori the number of movement patterns to be learned.
This issue can be mitigated by adopting growing extensions of the SOM such as the~GWR~algorithm \cite{marsland2002self}. 
The GWR algorithm has the advantage of a nonfixed, but varying topology that requires no specification of the number of neurons in advance.
Moreover, the prediction capability of the self-organizing approaches in the case of multidimensional data sequences has not been thoroughly analyzed in the literature. 
In the current work, we present experimental results in the context of a challenging robotic task, whereby real-world sensorimotor sequences have to be learned and predicted. 
\begin{figure*}[ht]
      \centering
      \includegraphics[scale=.53]{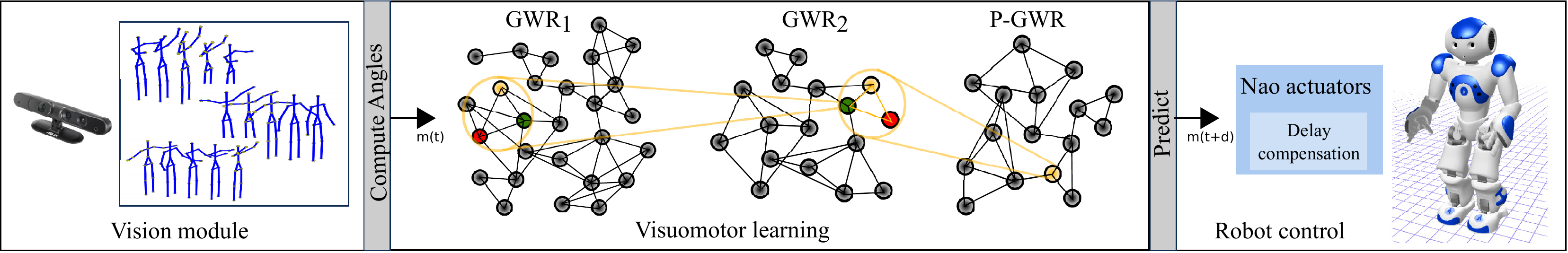}      
      \caption{Overview of the proposed system for the sensorimotor delay compensation during an imitation scenario. The vision module acquires motion from a depth sensor and estimates the three-dimensional position of joints. Shoulder and elbow angle values are extracted and fed to the visuomotor learning algorithm. The robot then receives predicted motor commands processed by the delay compensation module.}
      \label{fig:architecture}
\end{figure*} 

\subsection{Incremental learning of motion patterns}
In the context of learning motion sequences, an architecture capable of incremental learning should identify unknown patterns and adapt its internal structure in consequence.  
This topic has been the focus of a number of studies on \textit{programming by demonstration }(PbD)~\cite{Billard2016}.
Kuli\'c \textit{et al.}~\cite{kulic2008incremental}~used HMMs for segmenting and representing motion patterns together with a clustering algorithm that learns in an incremental fashion based on intra-model distances.
In a more recent approach, the authors organized motion patterns as leaves of a directed graph where edges represented temporal transitions \cite{kulic2012incremental}.
However, the approach was built upon automatic segmentation which required observing the complete demonstrated task, thereby becoming task-dependent. 
A number of other works have also adapted HMMs to the problem of incremental learning of human motion~\cite{takano2006humanoid}\cite{billard2006discriminative}\cite{ekvall2006online}\cite{dixon2004predictive}. The main drawback of these methods is their requirement for knowing a priori the number of motions to be learned or the number of Markov models comprising the learning architecture. 

Ogata et al.~\cite{ogata2004open}~proposed a model that considers the case of long-term incremental learning. 
In their work, a recurrent neural network was used to learn a navigation task in cooperation with a human partner. 
The authors introduced a new training method for the recursive neural network in order to avoid the problem of memory corruption during new training data acquisition. 
Calinon~\textit{et al.}~\cite{calinon2007incremental} showed that the Gaussian Mixture Regression (GMR) technique can be successfully applied for encoding demonstrated motion patterns incrementally through a Gaussian Mixture Model (GMM) tuned with an expectation-maximization (EM) algorithm.
The main limitation of this method is the need to specify in advance the number and complexity of tasks in order to find an optimal number of Gaussian components.
Therefore, Khansari-Zadeh and Billard~\cite{khansari2010bm} suggested a learning procedure capable of modelling demonstrated motion sequences through an adaptive GMM. 
Cederborg~\textit{et al.}~\cite{cederborg2010incremental} suggested to perform a local partitioning of the input space through kd-trees and training several local GMR models.

However, for high-dimensional data, partitioning of input space in a real-time system requires additional computational time.
Regarding this issue, it is convenient to adopt self-organized network-based methods that perform in parallel partitioning of the input space through the creation of prototypical representations as well as the fitting of necessary local models.  
The application of a growing self-organizing network, such as the GWR, allows for the learning of prototypical motion patterns in an incremental fashion~\cite{PMW16}. 

\section{Methodology}
\subsection{Overview}
The proposed learning architecture consists of a hierarchy of GWR networks~\cite{marsland2002self} which process input data sequences and learn inherent spatiotemporal dependencies~(Fig.~\ref{fig:architecture}). 
The first layer of the hierarchy learns a set of spatial prototype vectors which will then encode incoming data samples. 
The temporal dependence of the input data is captured as temporally ordered concatenations of consecutively matched prototypes which become more complex and of higher dimensionality when moving towards the last layer. 
When body motion sequences are provided, the response of the neurons in the architecture resembles the neural selectivity towards temporally ordered body pose snapshots in the human brain~\cite{giese2003neural}. 
This simple, but effective data sequence representation is also convenient in a prediction application due to implicitly mapping past values to the future ones. 
The concatenation vector is composed of two parts: the first part carries information about the input data at previous time steps and the second part concerns the desired output of this mapping. 

The evaluation of the predictive capabilities of the proposed architecture for compensating robot sensorimotor delay will be conducted in an imitation scenario where a simulated Nao robot imitates a human demonstrator while compensating for the sensorimotor delay in an on-line manner. 

\subsection{Learning with the GWR algorithm}\label{subsec:gwr}
The GWR network is composed of neurons and edges that link the neurons forming neighborhood relationships.
The network starts with a set of two neurons randomly initialized and, during the learning iterations, both neurons and edges can be created, updated, or removed.
At each learning iteration, $t$,  the first and the second best-matching units (BMUs) are computed as the neurons with the smallest Euclidean distance with the input sample $\textbf{x}(t)$.
The activity of the network, $a(t)$, is computed as a function of the Euclidean distance between the weight vector of the first BMU, $\textbf{w}_b$, and the input data sample $\textbf{x}(t)$:
\begin{equation}\label{eq:activation}
a = \exp ( { - ||\textbf{x}(t) - \textbf{w}_b|| } ).
\end{equation}
Whenever the activity of the network is smaller than a given threshold $a_T$, a new neuron is added with a weight vector:
\begin{equation}
\textbf{w}_r = 0.5 \cdot (\textbf{x}(t) + \textbf{w}_b)
\end{equation}
The activation threshold parameter, $a_T$, modulates the amount of generalization, i.e., the largest discrepancy between an incoming stimulus and its BMU.
Edges are created between the first and the second BMUs.
An edge ageing mechanism takes care of removing rarely activated edges, i.e., edges exceeding the age threshold, and unconnected neurons consequently.
In this way, representations of data samples that have been seen in the far past are eliminated leading to an efficient use of available resources from the lifelong learning perspective.
Moreover, a firing rate mechanism that measures how often each neuron has been activated by the input leads to a sufficient training before new neurons are created. 
The firing rate is initially set to one and than decreases every time a neuron and its neighbors are activated in the following way:
\begin{equation}\label{eq:hab}
\Delta h_i = \rho_i \cdot \kappa \cdot (1 - h_i) - \rho_i,
\end{equation}
where $\rho_i$ and $\kappa$ are the constants controlling the behaviour of the decreasing function curve.
Typically, the $\rho$ constant is set higher for the BMU ($\rho_b$) than for its topological neighbors ($\rho_n$).
Given an input data sample $\textbf{x}(t)$, if no new neurons are added, the weights of the first BMU and its neighbors are updated as follows:
\begin{equation}\label{eq:update}
\Delta\textbf{w}_i = \epsilon_i \cdot h_i \cdot (\textbf{x}(t) - \textbf{w}_i),
\end{equation}
\noindent where $\epsilon_i$ and $h_i$ are the constant learning rate and the firing counter variable respectively. 
The learning of the GWR algorithm stops when a given criterion is met, e.g., a maximum network size or a maximum number of learning epochs.

\subsection{Temporal sequence representations}\label{subsection:sequence_representation}
GWR networks do not encode temporal relationships of the input.
This limitation has been addressed by different extensions, such as hierarchies of GWRs augmented with a \textit{window in time} memory or recurrent connections~\cite{parisiFrontiers}\cite{Parisi2016}\cite{Mici2016}. Since our goal is to both encode data sequences and generate them, we adopt the first approach, in which the relevant information regarding data samples in a window of time is always explicitly available. 
Moreover, the use of a hierarchy instead of a shallow architecture allows for the encoding of multiple time-varying sequences through prototype neurons which can be reused for representing different sequences.

\begin{figure}
      \centering
      \includegraphics[scale=.5]{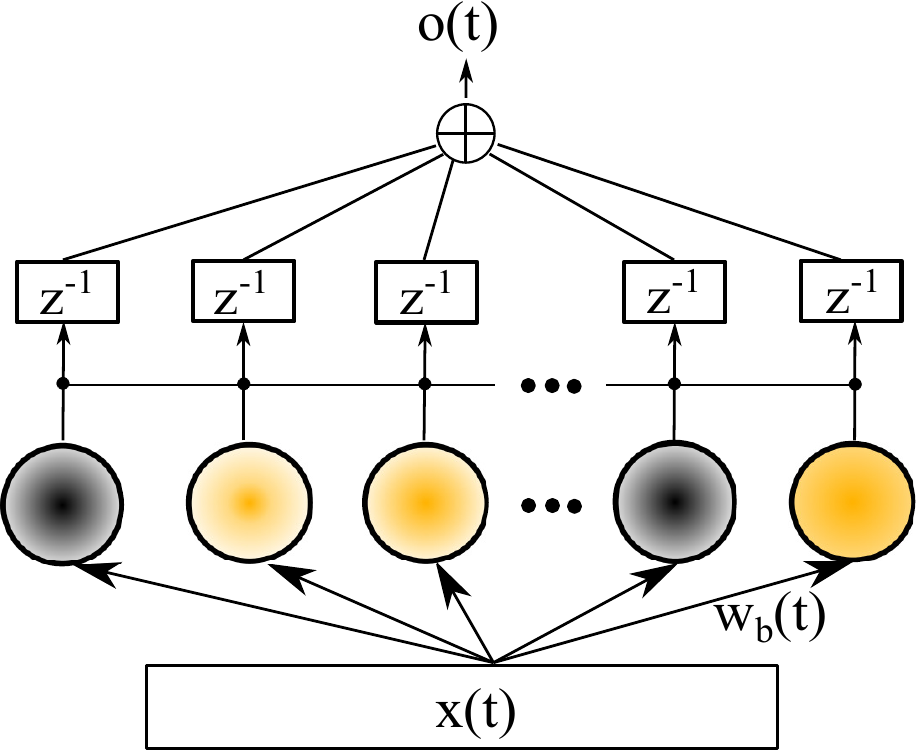}
      \caption{Schematic description of the output computation for the first two layers of our learning architecture (not all neurons and connections are shown).~Given an input data sample $\textbf{x}(t)$, the weight of the best-matching unit is concatenated with the weights of the previously activated neurons (depicted in fading yellow) in order to compute the output $\textbf{o}(t)$. The length of the concatenation vector is a pre-defined constant $\tau$~($\tau=3$ in this example). The $z^{-1}$ blocks denote the time delay.}
      \label{fig:time_window}
\end{figure}

The GWR learning mechanism described in Section~\ref{subsec:gwr} is employed for training the first two layers of the proposed architecture, namely the \textit{GWR}$_1$ and \textit{GWR}$_2$ networks.
The output of both networks is computed as the concatenation of the weights of consecutively activated neurons within a pre-defined temporal window $\tau$~(see Fig.\ref{fig:time_window}):
\begin{equation}\label{eq:output_function}
\textbf{o}(t) = \textbf{w}_{b(t)} \oplus \textbf{w}_{b(t-1)} \oplus ... \oplus \textbf{w}_{b(t - \tau + 1)},
\end{equation}  
\noindent where $\oplus$ denotes the concatenation operation.
Moving up in the hierarchy, the output $\textbf{o}(t)$ will represent the input for the GWR network of the higher layer. 
In this way, the \textit{GWR}$_1$ network learns a dictionary of prototypes of the spatial body configurations domain, while the \textit{GWR}$_2$ and \textit{P-GWR} networks encode body motion patterns accumulated over a short and a longer time period respectively. 

Following this hierarchical learning scheme, we adapt the GWR neuron elimination strategy in a layer-wise manner to address the problem of forgetting rarely encountered, but still relevant information. 
For instance, at the level of the \textit{GWR}$_1$ network, which represents spatial body configurations, it is more probable that rarely seen input data samples are due to sensory noise.
Therefore, we can set a lower edge age threshold here, leading to a higher rate of neuron elimination.
For the \textit{GWR}$_2$ and \textit{P-GWR} networks, on the other hand, rarely seen data samples are most probably due to sub-sequences encountered in the far past. 
We can set a higher edge age threshold so that neurons are removed more rarely.

\subsection{The Predictive GWR algorithm}
The problem of one-step-ahead prediction can be formalized as a function approximation problem. 
Given a multi-dimensional time series denoted by \{\textbf{y}(t)\}, the function approximation is of the form:
\begin{equation}
\hat{\textbf{y}}(t+1) = \hat{f}\left(\textbf{y}(t), \textbf{y}(t-1),..., \textbf{y}(t-(p-1)) | \Theta \right),
\end{equation}
\noindent where the input of the function, or \textit{regressor}, has an order of regression $p \in \mathbb{Z}^+$, with $\Theta$ denoting the vector of adjustable parameters of the model and $\textbf{y}(t+1)$ is the predicted value.
In other words, the prediction function maps the past $p$ input values to the observed value $\textbf{y}(t+1)$ directly following them. 
We adapt the GWR learning algorithm in order to implement this input-output mapping and apply this learning algorithm to the last layer of our architecture, i.e., to the \textit{P-GWR} network.

The input samples fed to the \textit{P-GWR} network are concatenations of the temporally ordered BMUs from the preceding layer~(Eq.~\ref{eq:output_function}).
We divide the input into two parts: the regressor, $\textbf{x}^{in}(t)$, and the desired output, i.e., the value to predict $\textbf{x}^{out}(t)$:
\begin{equation}
\begin{split}
\textbf{x}^{in}(t) = \textbf{x}(t) \oplus \textbf{x}(t-1) \oplus ... \oplus \textbf{x}(t-p+1), \\
\textbf{x}^{out}(t) = \textbf{x}(t+1), 
\end{split}
\end{equation}
\noindent with $p$ denoting the maximum index of the past values.
Each neuron of the \textit{P-GWR} network will then have two weight vectors which we will call the input $\textbf{w}^{in}$ and the output $\textbf{w}^{out}$ weight vectors.
During training, the input weight vector will learn to represent the input data regressor and the output weight vector will represent the corresponding predicted value. 
This learning scheme has been successfully applied to the \textit{Vector-Quantized Temporal Associative Memory} (VQTAM) model~\cite{barreto2007time}, shown to perform well on tasks such as time series prediction and predictive control~\cite{barreto2003self}. 

The learning procedure for the Predictive GWR algorithm resembles the original GWR with a set of adaptations for temporal processing. 
During training, the first and the second best-matching units,~$b$ and $s$, at time step $t$ are computed considering only the regressor part of the input:
\begin{equation}\label{eq:bmu}
\begin{split}
b = arg \min\limits_{n \in A} ||\textbf{x}^{in}(t) - \textbf{w}^{in}_n||, \\
s = arg \min\limits_{n \in A / \{b\}} ||\textbf{x}^{in}(t) - \textbf{w}^{in}_n||,
\end{split}
\end{equation}
where $\textbf{w}^{in}_n$ is the input weight vector of the neuron $n$ and $A$ is the set of all neurons.
However, for the weight updates both $\textbf{x}^{in}(t)$ and $\textbf{x}^{out}(t)$ are considered: 
\begin{equation}\label{eq:dual_weight_update}
\begin{split}
\Delta \textbf{w}^{in}_i = \epsilon_i \cdot c_i \cdot (\textbf{x}^{in}(t) - \textbf{w}^{in}_i), \\
\Delta \textbf{w}^{out}_i = \epsilon_i \cdot c_i \cdot (\textbf{x}^{out}(t) - \textbf{w}^{out}_i),
\\
\end{split}
\end{equation}
with the learning rates $0 < \epsilon_i < 1$ being higher for the BMUs ($\epsilon_b$) than for the topological neighbors ($\epsilon_n$).	
This learning mechanism guaranties that the regressor space is vector-quantized while the prediction error is minimized at each learning iteration.

The Predictive GWR algorithm operates differently from supervised prediction approaches. In the latter, the prediction error signal is the factor that guides the learning, whereas in the Predictive GWR the prediction error is implicitly computed and minimized without affecting the learning dynamics. 
Moreover, unlike the SOM-based VQTAM model, the number of input-output mapping neurons, or \textit{local models}, is not pre-defined nor fixed, but instead adapts to the input data.
It should be noted that overfitting does not occur with the growth of the network due to the fact that neural growth decreases the quantization error which is proportional to the prediction error.

\subsection{Predicting sequences} \label{subsection:prediction}
Given an input regressor at time step $t$,~$\textbf{x}^{in}(t)$,~the one-step-ahead estimate is defined as the output weight vector of the \textit{P-GWR} best-matching unit:
\begin{equation}\label{eq:prediction_function}
\hat{\textbf{y}}(t+1) = \textbf{w}^{out}_{b}
\end{equation}
\noindent where $b$ is the index of the best-matching unit~(Eq.~\ref{eq:bmu}). 
In the case that the desired prediction horizon is greater than $1$, the multi-step-ahead prediction can be obtained by feeding back the predicted values into the regressor and computing Eq.~\ref{eq:bmu} recursively until the whole desired prediction vector is obtained.
An alternative to the recursive prediction is the vector prediction which is obtained by increasing the dimension of the $\textbf{x}^{out}$ vector with as many time steps as the desired prediction horizon $h$. 
Thus, the input regressor and the desired output would have the following form:

\begin{equation}
\begin{split}
\textbf{x}^{in}(t) = \textbf{x}(t) \oplus \textbf{x}(t-1) \oplus ... \oplus \textbf{x}(t-p+1), \\
\textbf{x}^{out}(t) = \textbf{x}(t+1) \oplus \textbf{x}(t+2) \oplus ... \oplus \textbf{x}(t+h),
\end{split}
\end{equation}
\noindent where $p$ denotes the index of the past values. 
The same dimensionality should be defined for the weight vectors $\textbf{w}^{in}$ and $\textbf{w}^{out}$ of the  \textit{P-GWR} neurons as well. 
This solution requires the training of the architecture with this setting of the weights.   

\section{The imitation scenario}

\subsection{Overview}
The scenario consists of a Nao robot incrementally learning  a set of visually demonstrated body motion patterns and directly imitating them while compensating for the sensorimotor delay. 
We showcase the predictive capabilities of the proposed architecture in the context of an imitation scenario motivated by the fact that it can potentially imply behaviour synchronization in the human-robot interaction. 
For humans, the synchronization of behavior is a fundamental principle for motor coordination and is known to increase rapport in daily social interaction~\cite{lorenz2011synchronization}. 
Psychological studies have shown that during conversation humans tend to coordinate body posture and gaze direction~\cite{shockley2009conversation}. 
This phenomenon is believed to be connected to the mirror neuron system~\cite{tessitore2010motor}, suggesting a common neural mechanism for both motor control and action understanding. 
Interpersonal coordination is an integral part of human interaction, thus we assume that applied to HRI scenarios it may promote the social acceptance of robots. 

A schematic description of the proposed system is given in Fig.~\ref{fig:architecture}.
The user's body motion is the input into the model and the motor commands for the robot are obtained by mapping the user's arm skeletal structure to the robot's arm joint angles.
This direct motion transfer allows for a simple, yet compact representation of the visuomotor states that does not require the application of computationally expensive inverse kinematics algorithms.
Demonstrated motion trajectories are learned incrementally by training our hierarchical GWR-based learning algorithm.
This allows for extracting prototypical motion patterns which can be used for the generation of robot movements as well as prediction of future target trajectories in parallel.
In this robot task, the prediction of future visuomotor states is necessary to compensate for the sensory delay introduced by the vision sensor camera, the signal transmission delay as well as the robot's motor latency during motion generation. 
The simulated Nao robot is used as the robotic platform for the experimental evaluation.

\begin{figure}
\centering
    \begin{subfigure}[b]{\linewidth}
        \includegraphics[width=0.95\textwidth]{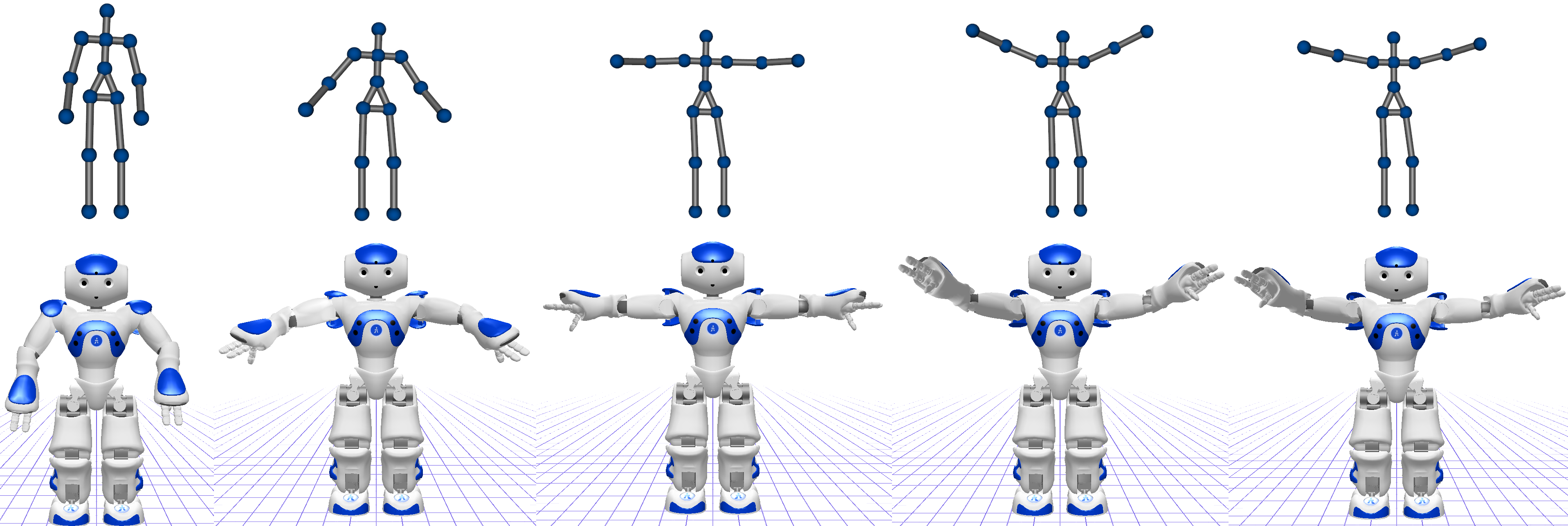}
    \end{subfigure}
    \begin{subfigure}[b]{\linewidth}
        \includegraphics[width=0.95\textwidth]{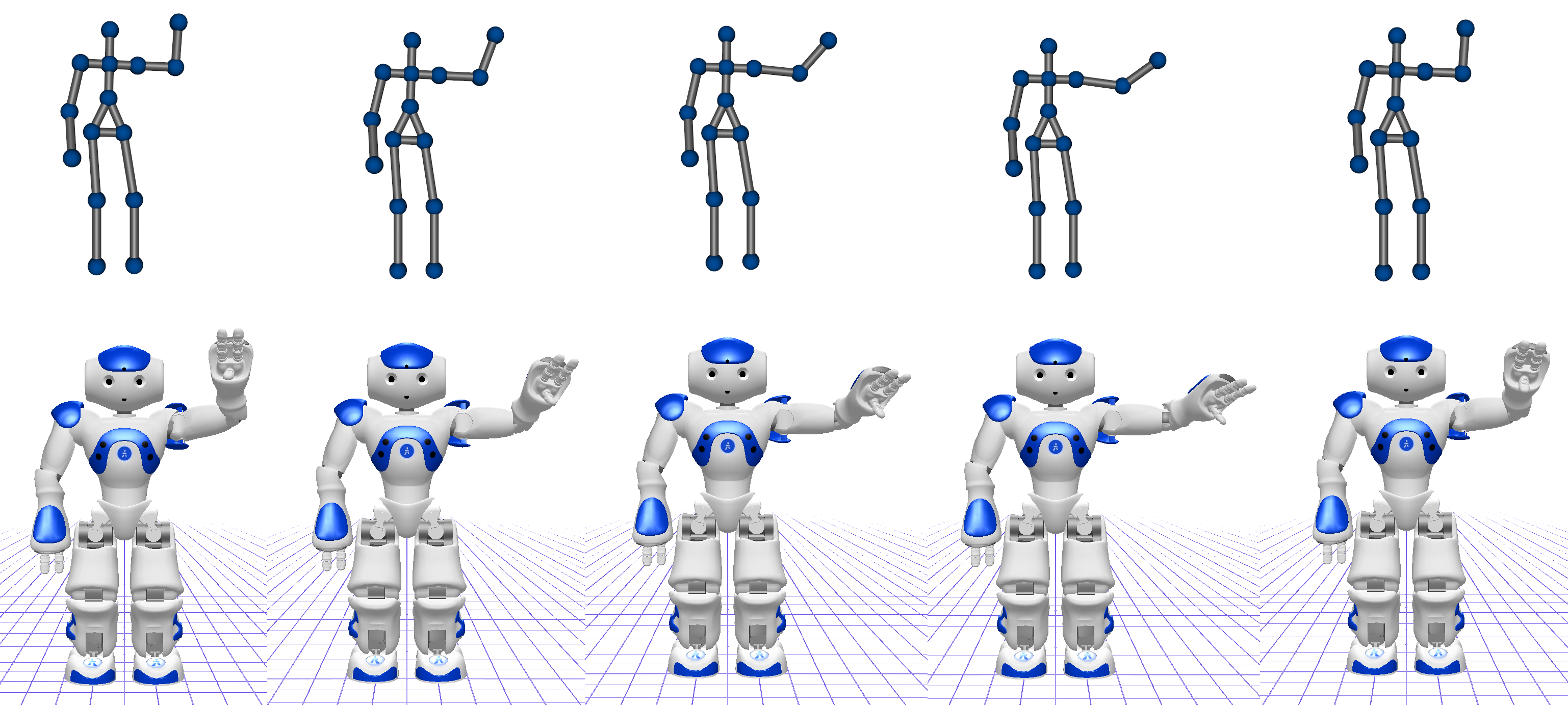}
    \end{subfigure}
    \begin{subfigure}[b]{\linewidth}
        \includegraphics[width=0.95\textwidth]{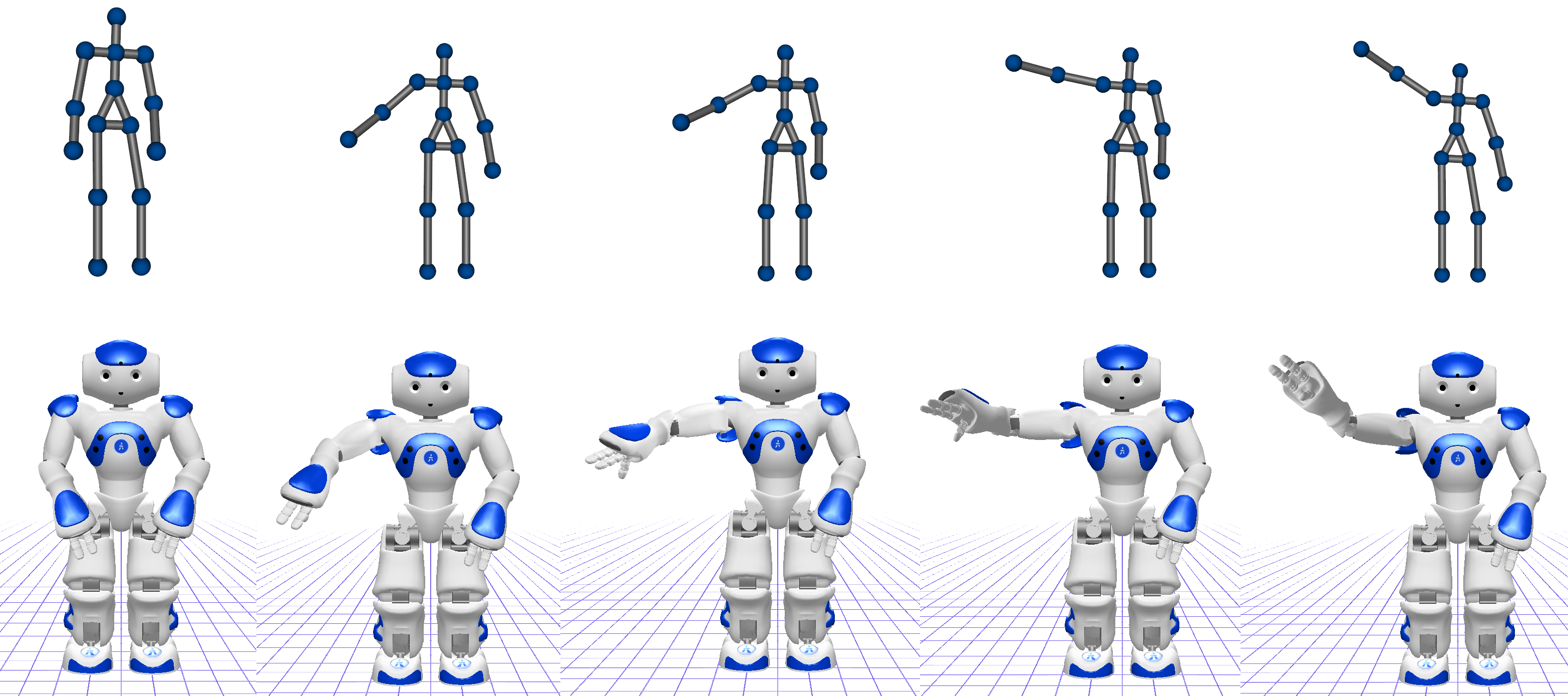}
    \end{subfigure}
\label{fig:robot}
\caption{Examples of arm movement patterns. The visual input data, represented as three-dimensional skeleton sequences, are mapped to the robots' joint angles.}
\end{figure}
 
\subsection{System description}
A general overview of the proposed architecture is depicted in Fig.~\ref{fig:architecture}. The system consists of three main modules: 
\begin{enumerate*}
\item The \textit{vision module}, which includes the depth sensor and the tracking of the 3D skeleton through OpenNI/NITE framework;~\footnote{OpenNI/NITE:~\href{
http://www.openni.org/software}{http://www.openni.org/software}}
\item The \textit{visuomotor learning} module, which receives angle values and provides future motor commands;
\item The \textit{robot control} module, which processes motor commands and relays them to the microcontrollers of the robot, which in our case is a locally simulated Nao. 
\end{enumerate*}

Our contribution is the visuomotor learning module which performs incremental adaptation and early prediction of human motion patterns.
Although the current setup uses a simulated environment, we will consider a further extension of the experiments towards the real robot.
Therefore, we simulate the same amount of motor response latency as it has been quantified in the real Nao robot, being between $30$ to $40$ ms \cite{Zhong2012}.
This latency could be even higher due to reduced motor performance, friction or weary hardware. 
Visual sensor latency on the other hand, for an RGB and depth resolution of 640x480, together with the computation time required from the skeleton estimation middleware can peak up to $500$~ms~\cite{livingston2012performance}.
Taking into consideration also possible transmission delays due to connectivity issues, we assume a maximum of $600$~ms of overall sensorimotor latency in order to carry out experiments described in Section~\ref{sec:experimental_results}.

\begin{figure*}
\centering
	\begin{subfigure}[b]{\linewidth}
		\centering		
        \quad \includegraphics[scale=.09]{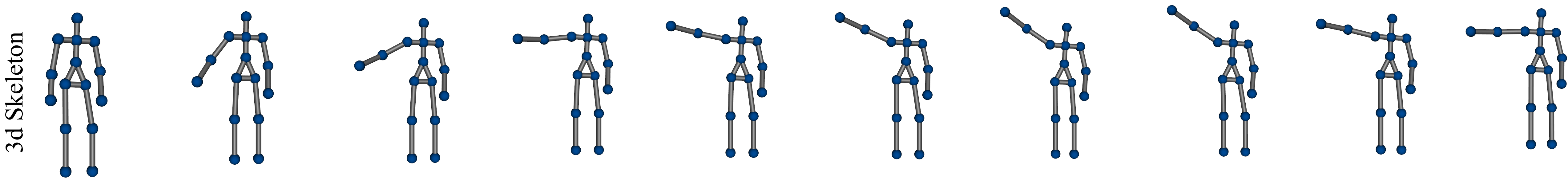}
    \vspace{.8em}
    \end{subfigure}
    \begin{subfigure}[b]{\linewidth}
    \centering
        \includegraphics[width=0.95\textwidth]{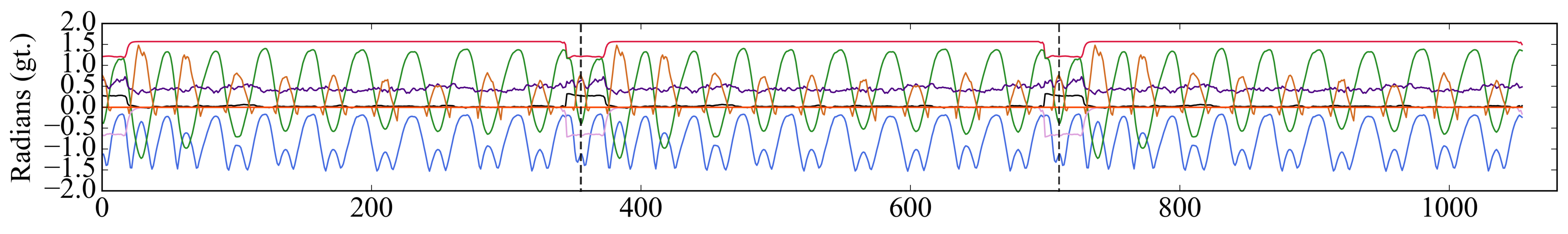}
    \end{subfigure}
    \begin{subfigure}[b]{\linewidth}
    \centering
        \includegraphics[width=0.95\textwidth]{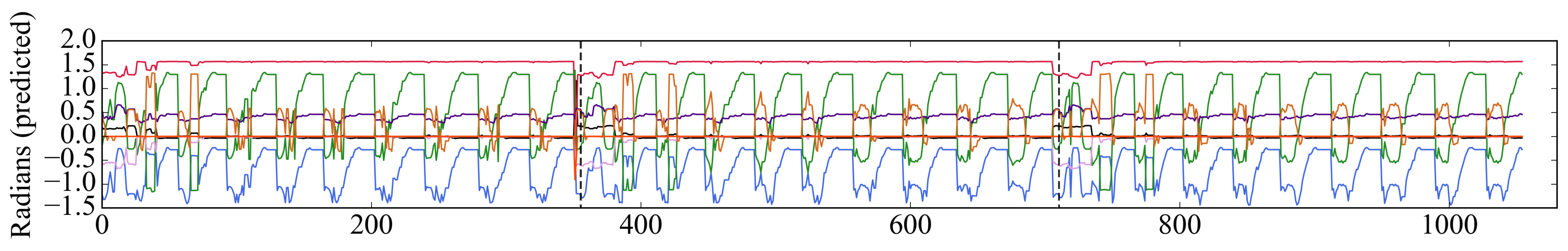}
    \end{subfigure}
    \begin{subfigure}[b]{\linewidth}
    \centering
        \includegraphics[width=0.975\textwidth]{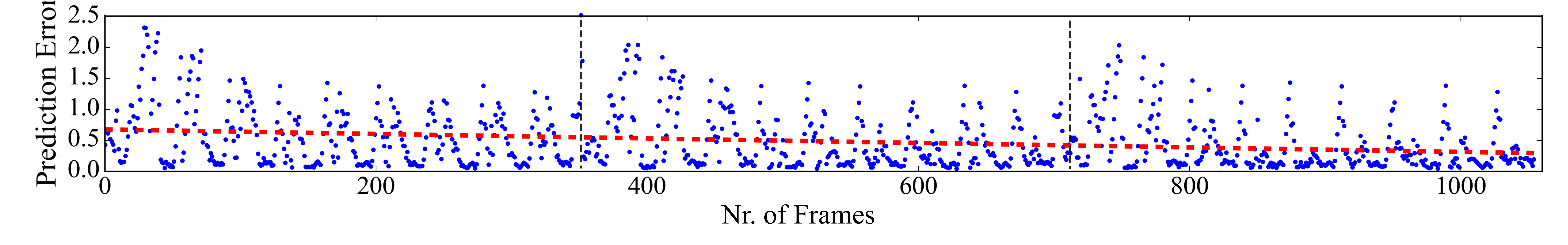}
    \end{subfigure}
            \caption{Behavior of the proposed architecture during training on an \textit{unseen} sequence demonstrated by one subject (the sequence is presented three times to the network). From top to bottom illustrated are: the skeleton model of the visual sequence, the ground truth data of robot joint angles, the values predicted from the network, and the Euclidean distance between predicted values and the ground truth over time (red dashed line indicating the statistical trend).}
            \label{fig:online_response}
\end{figure*}

\subsection{Data acquisition and representation}\label{subsec:data_representation}
The motion sequences were collected with an Asus Xtion Pro camera at 30 frames per second.
This type of sensor is capable of providing synchronized color information and depth maps at a reduced power consumption and weight, making it a more suitable choice than a Microsoft Kinect for being placed on our small humanoid robot.
Moreover, it offers a reliable and markerless body tracking method \cite{HanMicrosoft} which makes the interface less invasive.
The distance of each participant from the visual sensor was maintained between the sensor's operational range, i.e., $0.8-3.5$ meters.
To attenuate noise, we computed the median value for each body joint every 3 frames resulting in 10 joint position vectors per second~\cite{Parisi2016}.

We selected joint angles to represent the demonstrator's postures.
Joint angles allow a straightforward reconstruction of the regressed motion without applying inverse kinematics, which may be difficult due to redundancy and leads to less natural movements.
Nao's arm kinematic configuration differs from the human arm in terms of degrees of freedom (DoF).
For instance, the shoulder and the elbow joints have only two DoFs while human arms have three.
For this reason, we compute only shoulder \textit{pitch} and \textit{yaw} and elbow \textit{yaw} and \textit{roll} from the skeletal representation by applying trigonometric functions and map them to the Nao's joints by appropriate rotation of the coordinate frames.
Wrist orientations are not considered since they are not provided by the OpenNI/NITE framework. 
Considering the two arms, a frame contains a total of 8 angle values of body motion, which are given as input to the visuomotor learning module.
 
\section{Experimental Results}\label{sec:experimental_results}
We conducted experiments with a set of movement patterns that were demonstrated either with one or with both arms simultaneously: raise arm(s) laterally,  raise arm(s) in front, wave arm(s), rotate arms in front of the body both clockwise and counter-clockwise. Some of the movement patterns are illustrated in Fig.~3. 
In total, 10 different motion patterns were obtained, each repeated 10 times by three participants~(one female and two male) who were given no explicit indication of the purpose of the study nor instructions on how to perform the arm movements.
In total, we obtained 30 demonstrations for each of the pattern.
We first describe the incremental training procedure, then we assess and analyze in details the prediction accuracy of the proposed learning method.
We focus on the learning capabilities of the method while simulating a possible recurring malfunctioning of the visual system leading to loss of entire data chunks.
We conclude with a model for choosing the optimal predicted value for a system with a variable delay. 

\subsection{Hierarchical training}
The training of our architecture is carried out in an on-line manner. 
This requires that the GWR networks are trained sequentially one data sample at a time.
The initialization phase sees all networks composed of two neurons with random weight vectors, i.e., carrying no relevant information about the input data. 
The \textit{GWR}$_1$ network is trained in order to perform spatial vector quantization. 
Then the current sequence is gradually encoded as a trajectory of activated neurons as described in Eq.~\ref{eq:output_function} and given in input to the \textit{GWR}$_2$ network of the second layer. 
The same procedure is then repeated for the second layer until the training of the full architecture is performed. 
The learning of the 30 demonstrations of one motion pattern from all three subjects constitutes an \textit{epoch}.

\begin{table}
\caption{Training parameters for each GWR network in our architecture for the incremental learning of sensorimotor patterns.}
\label{table:parameters}
\begin{center}
\begin{tabular}{l|c} 
Parameter & Value \\
\hline
Activation Threshold & $a_T = 0.98$\\
Firing Threshold & $f_T = 0.1$\\
Learning rates & $\epsilon_b = 0.1, \epsilon_n = 0.01$\\
Firing counter behavior & $\rho_b=0.3, \rho_n = 0.1,  \kappa = 1.05$\\
Maximum edge age & \{100,~200,~300\}\\
Training epochs & 50 \\
\hline
\end{tabular}
\end{center}
\end{table}

The learning parameters used throughout our experiments are listed in Table~\ref{table:parameters}. 
The parameters have been empirically fine-tuned by considering the learning factors of the GWR algorithm. 
The firing threshold $f_T$ and the parameters $\rho_b$, $\rho_n$, and $\kappa$ define the decrease function of the firing counter (Eq.~\ref{eq:hab}) and were set in order to have at least seven trainings of a best-matching unit before inserting a new neuron.
It has been shown that increasing the number of trainings per neuron does not affect the performance of a GWR network significantly~\cite{marsland2002self}.  
The learning rates are generally chosen to yield a faster training for the BMUs than for their topological neighbors. 
However, given that the neurons' decreasing firing counter modulates the weights update~(Eq.~\ref{eq:update}), an optimal choice of the learning rates has little impact on the architecture's behaviour in the long run. 
The training epochs were chosen by analysing the converging behaviour of the composing GWR networks in terms of neural growth.
     
The activation threshold parameter $a_T$, which modulates the number of neurons, has the largest impact on the architecture's behaviour.
The closer to $1$ this value is, the greater is the number of neurons created and the better is the data reconstruction during the prediction phase.
Therefore, we kept $a_T$ relatively high for all GWR networks.
We provide an analysis of the impact of this parameter on the prediction performance of our architecture in Section~\ref{sec:activation_threshold}. 
Finally, the maximum edge age parameter, which modulates the removal of rarely used neurons, was set increasingly higher with each layer.
As discussed in Section~\ref{subsection:sequence_representation}, the neurons activated less frequently in the lower layer may be representing noisy input data samples, whereas in higher layers the neurons capture spatiotemporal dependencies which may vary significantly from sequence to sequence.

\subsection{Predictive behavior} 
\begin{figure}
\centering
    \begin{subfigure}[b]{\linewidth}
        \includegraphics[width=\textwidth]{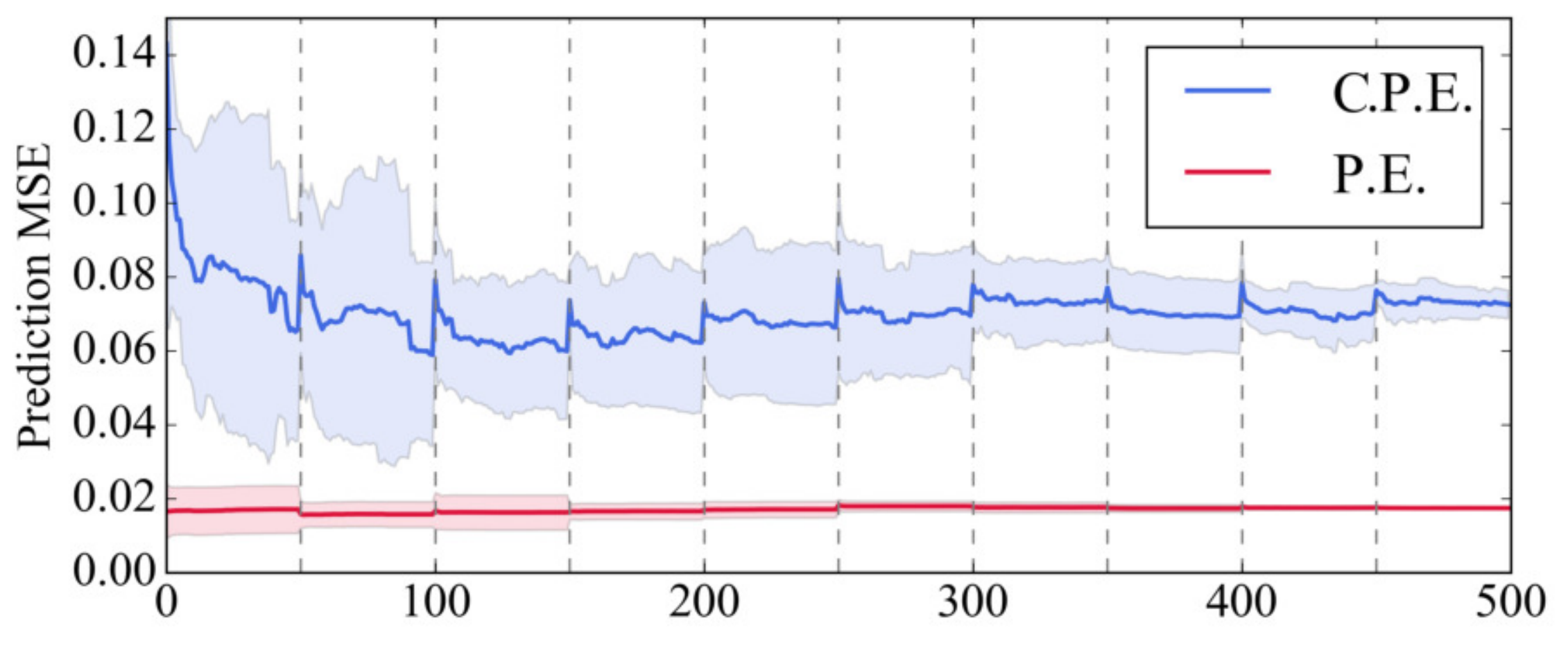}
        \caption{}
    \end{subfigure}
    \begin{subfigure}[b]{\linewidth}
        \includegraphics[width=\textwidth]{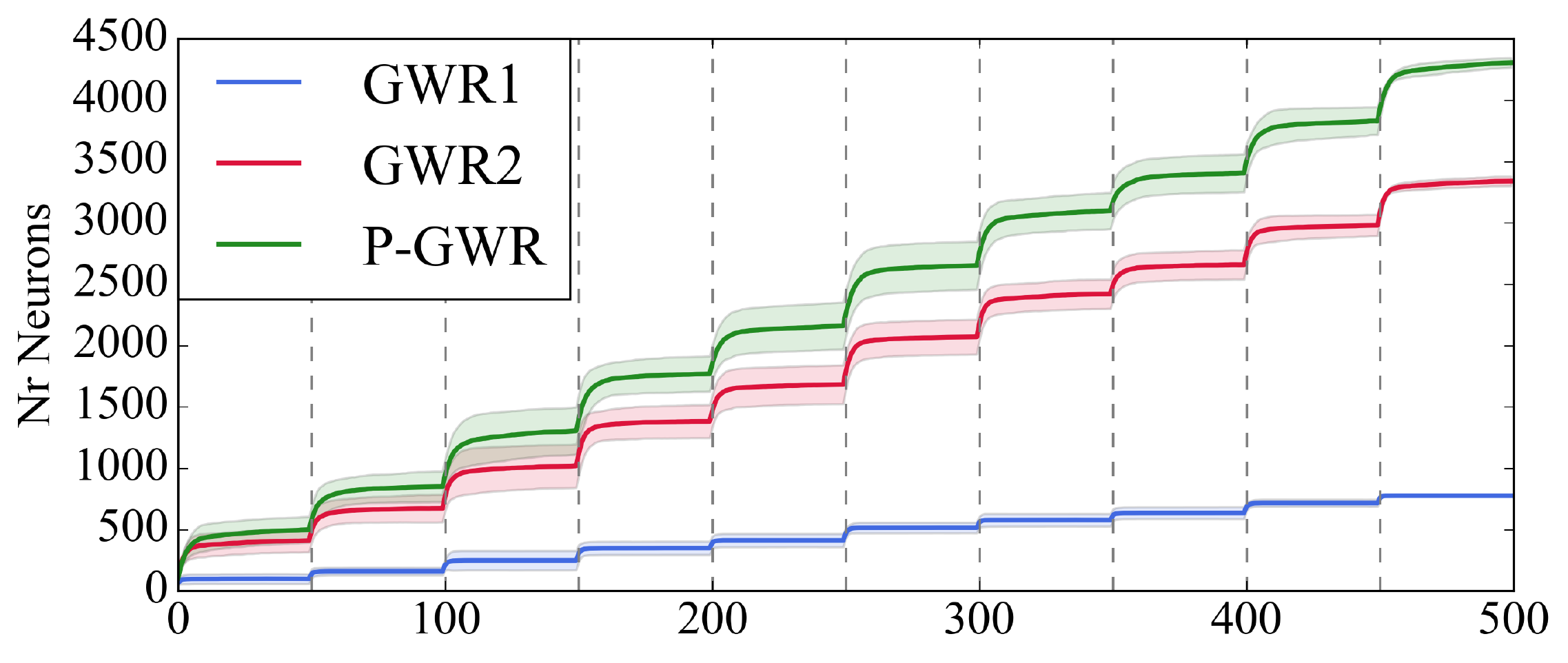}
        \caption{}
    \end{subfigure}
\label{fig:prediction}
\caption{(a)~The cumulative prediction error (C.P.E) averaged over all learned sequences up to each learning epoch (in blue) and the prediction error (P.E.) computed between the predicted sequence and the sequence represented by the architecture (in red), (b)~Average and standard deviation of the neural growth of the three GWR networks during learning.}
\end{figure}

We now assess the predictive capabilities of the proposed method while the training is occurring continuously. 
Considering that the data sample rate is 10 fps~(see Section~\ref{subsec:data_representation}), we set a prediction horizon of 6 frames in order to compensate for the estimated delay of 600 ms. 

\subsubsection{How fast does the architecture adapt to a new sequence?}
An example of the on-line response of the architecture is shown in Fig.~\ref{fig:online_response}. 
We observed that, except in cases of highly noisy trajectories, the network adapted to an unseen input already after a few video frames, e.g., $\approx100$ frames which correspond to $10$ seconds of the video sequence, and refined its internal representation after three presentations of the motion sequence demonstrated by one subject, i.e., after 30 demonstrations.
This can be seen by the statistical trend of the prediction error.

\subsubsection{Behaviour analysis and prediction performance during incremental learning}
We presented the movement sequences one at a time and let the architecture train for 50 epochs on each new sequence.
The training phase was in total of 500 epochs for the whole dataset.
Then, we re-ran the same experiment by varying the presentation order of the sequences and report the results averaged across all trials.
In this way, the behavior analysis does not depend on the order of the data given during training.
We analyzed the cumulative prediction error (C.P.E) of the model by computing the \textit{mean squared error} (MSE) over all movement sequences learned up to each training epoch.
For comparison, we also computed the MSE between the values predicted by the model and the sensory input after being processed by the \textit{GWR}$_1$ and the \textit{GWR}$_2$ networks. 
We refer to this performance measure as the prediction error (P.E.) since it evaluates directly the prediction accuracy of the \textit{P-GWR} network while removing the quantization error propagated from the first two layers.

The flow of the overall MSE during training and the neural growth of the GWR networks composing the architecture are reported in Fig.~5. 
The moment in which we introduce a new motion sequence is marked by a vertical dashed line. 
As expected, the cumulative prediction error increases as soon as a new sequence is introduced~(leading to the high peaks in Fig.~5.a.), for then decreasing immediately.
However, the error does not grow but stays constant even though new knowledge is being added every 50 learning epochs.
This is a desirable feature for an incremental learning approach.
In Fig.~b., we observe that with the introduction of a new motion sequence there is an immediate neural growth of the three GWR networks followed by the stabilisation of the number of neurons indicating a fast convergence.
This neural growth is an understandable consequence of the fact that the movement sequences are very different from each other.
In fact, the \textit{GWR}$_1$ network, performing quantization of the spatial domain, converges to a much lower number of neurons, whereas the higher layers, namely the \textit{GWR}$_2$ and the \textit{P-GWR} network, have to capture a high variance of spatiotemporal patterns.
However, the computational complexity of a prediction step is $O(n)$, where $n$ is the number of neurons. 
Thus the growth of the network does not introduce significant computational cost.

\subsubsection{Impact of the activation threshold}\label{sec:activation_threshold}
\begin{figure}
\centering
	\includegraphics[scale=.36]{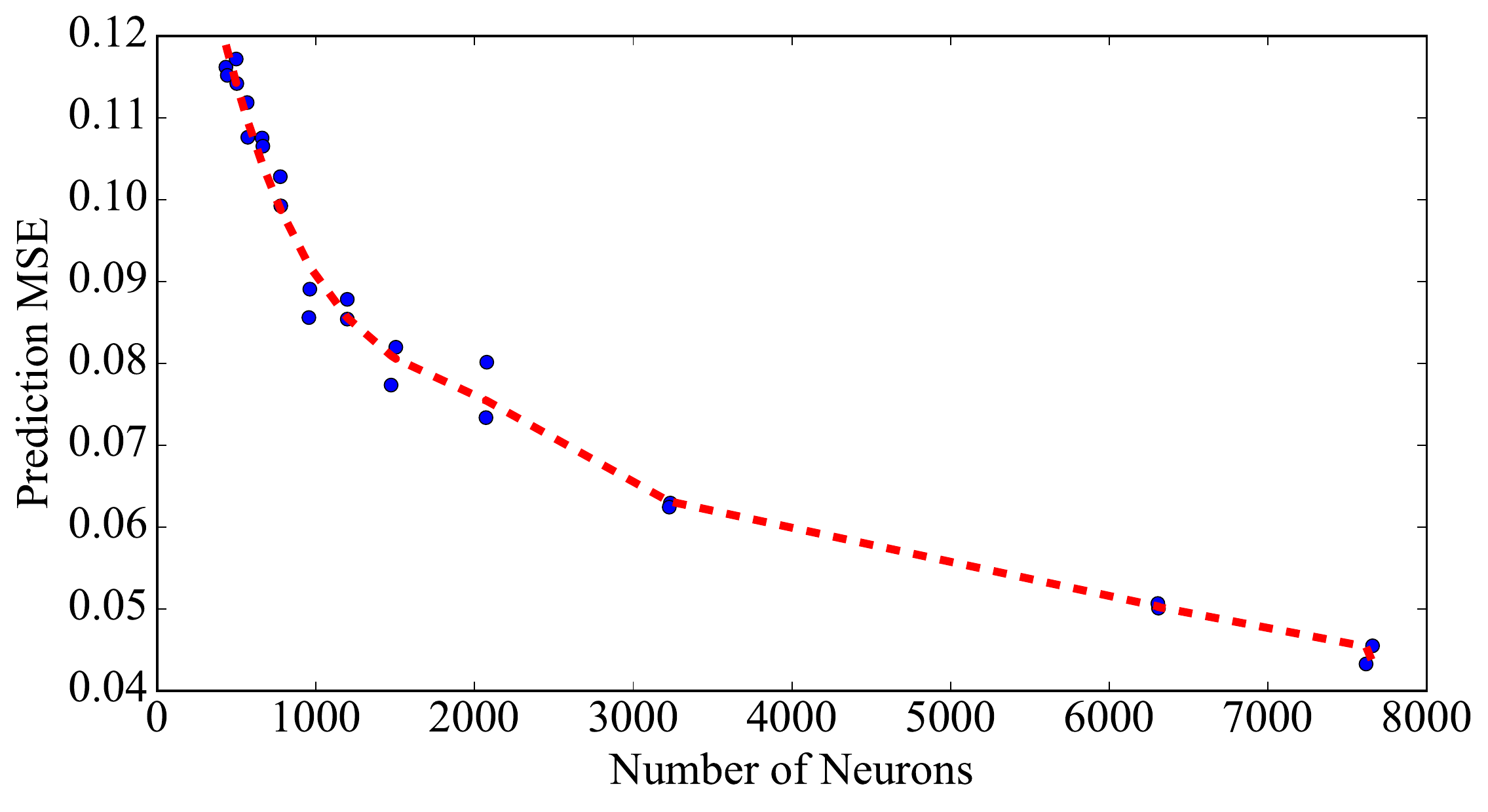}     
\caption{Prediction mean squared error (MSE) versus the number of neurons in the \textit{P-GWR} network.}    
\label{fig:neurons_vs_mse}	
\end{figure}

In the described experiments, we set a relatively high activation threshold parameter $a_T$ which led to a continuous growth of the GWR networks. 
Thus, we further investigated how a decreased number of neurons in the \textit{P-GWR} network would affect the overall prediction error.
For this purpose, we fixed the weight vectors of the first two layers after having been trained on the entire dataset, and ran multiple times the incremental learning procedure on the \textit{P-GWR} network, each time with a different activation threshold parameter $a_T \in \{0.5, 0.55, 0.6, ... ,0.9, 0.95, 0.99\}$. 
We observed that a lower number of neurons, obtained through lower threshold values, led to quite high values of the mean squared error~(Fig.~\ref{fig:neurons_vs_mse}). 
However, due to the hierarchical structure of our architecture, the quantization error can be propagated from layer to layer.
It is expected that similar performances can be reproduced with a lower number of neurons in the \textit{P-GWR} network when a lower quantization error is obtained in the preceding layers.  

\subsubsection{Sensitivity to the prediction horizon}
\begin{figure}
\centering
\includegraphics[width=.9\linewidth]{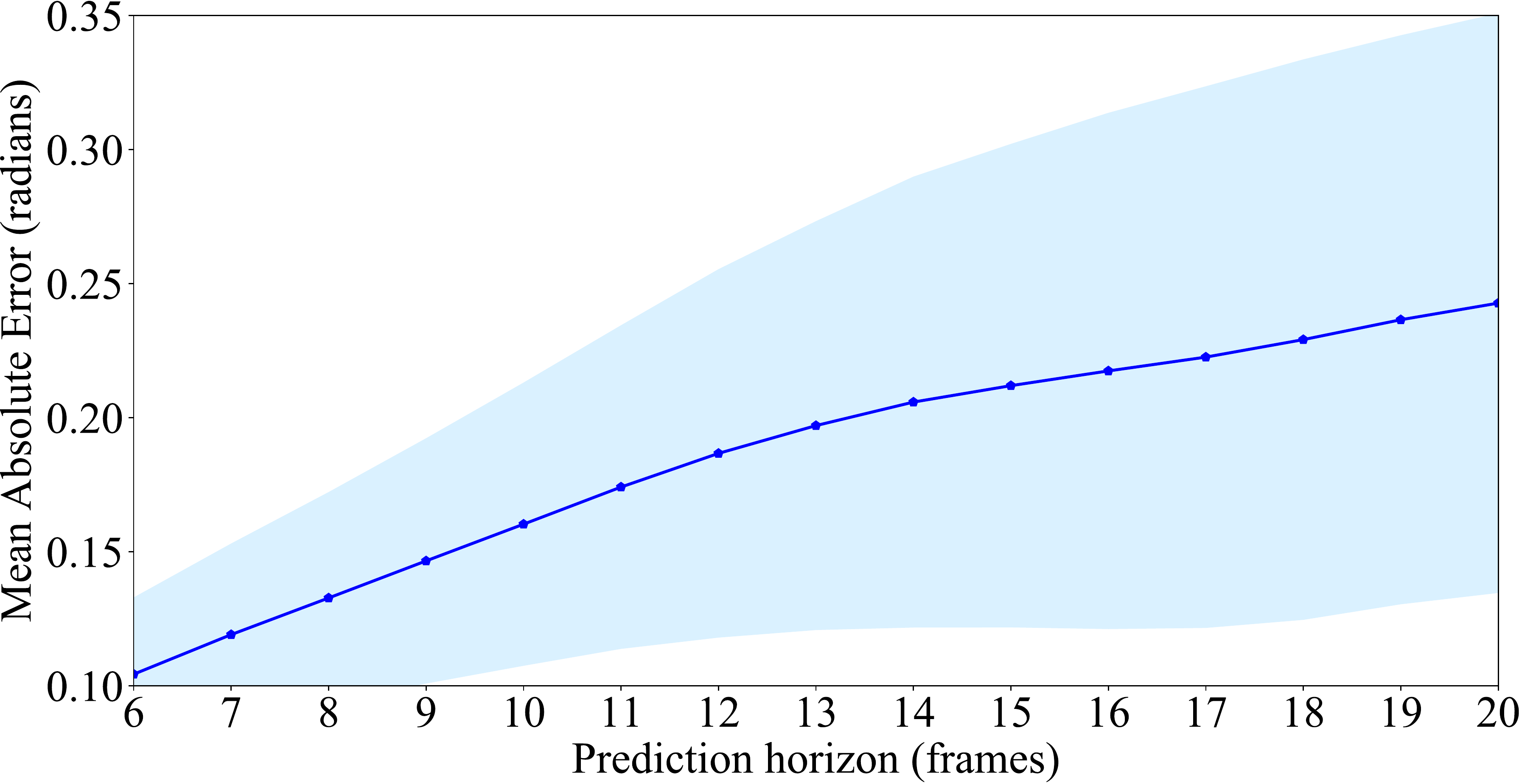}
\caption{Mean absolute error (in radians) for increasing values of prediction horizons (expressed in frames). In our case, 20 frames correspond to $2$ seconds of a video sequence.}
\label{fig:sensitivity}
\end{figure}

We now take the architecture trained on the whole dataset and evaluate its prediction accuracy while increasing the prediction horizon up to 20 frames which correspond to 2s of a video sequence.
For achieving multi-step-ahead prediction, we compute the predicted values recursively as described in Section~\ref{subsection:prediction}. 
In Fig.~\ref{fig:sensitivity}, we report the mean absolute error and the standard deviation in radians in order to give a better idea of the error range. 
The results show a relatively high magnitude of error for prediction horizons bigger than 10 frames.
This should come as no surprise since producing accurate long-term predictions is a challenging task when dealing with human-like motion sequences.
However, it seems that in average the error does not grow linearly but remains under 0.25 radians.

\subsection{Learning with missing sensory data}
\begin{figure}
\centering
\includegraphics[width=.9\linewidth]{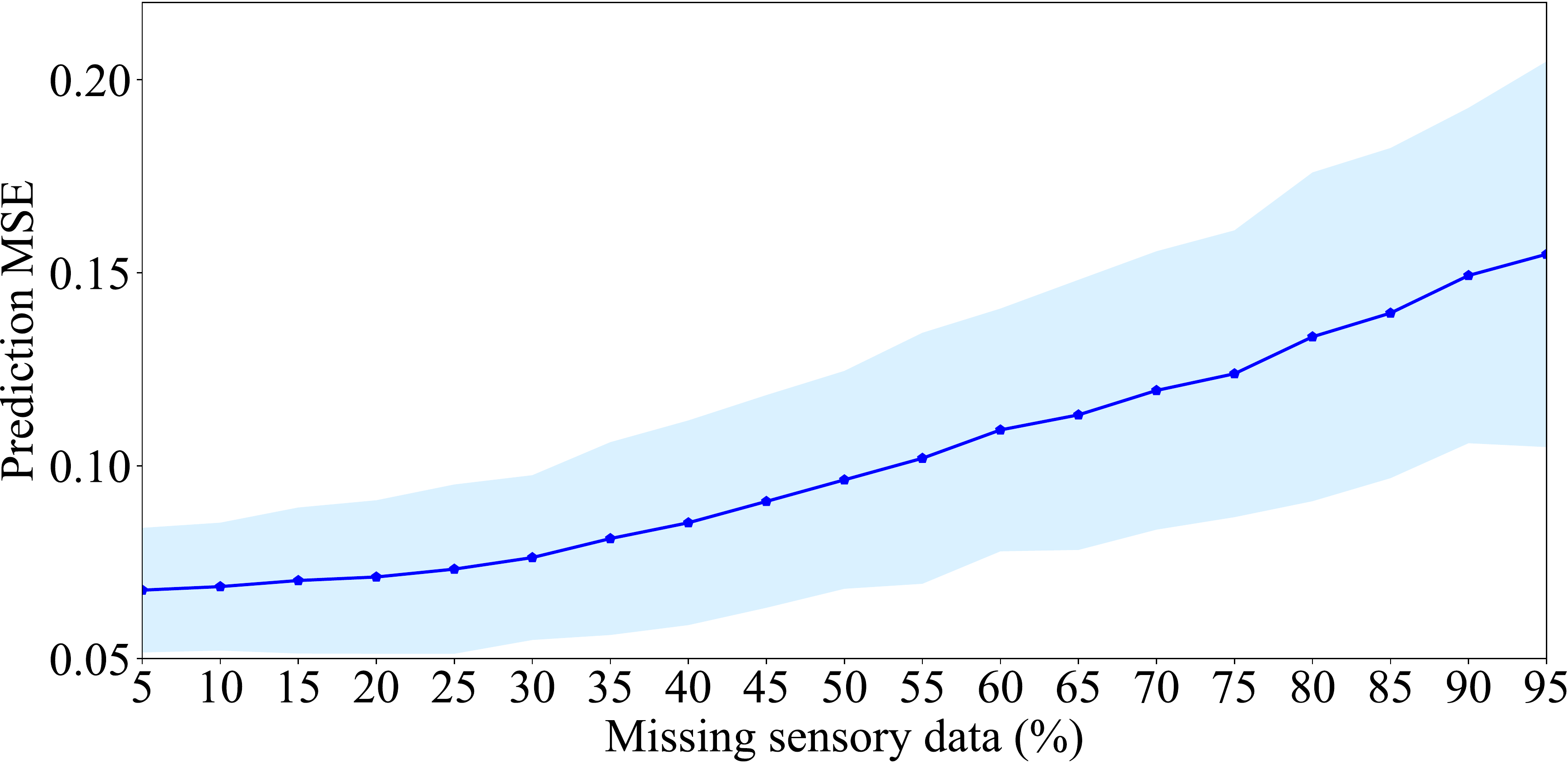}         
\caption{Prediction MSE averaged over 50 epochs of training on each motion pattern. For up to $30 \%$ of data loss the MSE does not grow linearly but rather stays almost constant. From this point on, the increasing percentage of data loss leads to an inevitable growth of the prediction error.}
\label{fig:data_loss}
\end{figure}

In the following set of experiments, we analyze how the predictive performance of the network changes when trained on input data produced by a faulty visual sensor.
We simulate an occurring loss of entire input data chunks in the following way: during the presentation of a motion pattern, we randomly choose video frames where a whole second of data samples (i.e., 10 frames) is eliminated.
The network is trained for 50 epochs on a motion sequence, each time with a different missing portion of information.
We repeat the experiment thereby increasing the occurrence of this event in order to compromise up to 95\% of the data and see how much the overall prediction error increases.
Results are averaged over epochs and are presented in Fig.~\ref{fig:data_loss}.
As it can be seen, the prediction MSE stays almost constant up to $30 \%$ of data loss.
This means that the network can still learn and predict motion sequences even under such circumstances.

\subsection{Compensating a variable delay}
Experimental results reported so far have accounted for compensating a fixed time delay which has been measured empirically by generating motor behavior with the real robot.
However, the proposed architecture can also be used when the delay varies due to changes in the status of the hardware.
In this case, given the configuration of the robot at time step $t$ in terms of joint angle values $J_{\xi}(t)$, where $\xi$ is the time delay estimation, the optimal predicted angle values to execute in the next step can be chosen in the following way:
$$
P^{*} = arg \min_{i \in [0, h]} ||J_{\xi}(t) - P(t+i)||, \eqno(11)
$$
\noindent where $P(t+i)$ are the predictions computed up to a maximum $h$ of the prediction horizon. 

The application of this prediction step requires a method for the estimation of the time-delay $\xi$, which is out of the scope of this work.
Current time-delay estimation techniques mainly cover constant time delays, random delay with a specific noise characteristic, or restricted dynamic time delay~\cite{sargolzaei2016sensorimotor}, which nonetheless do not address uncertainty affecting real-world robot applications.
Computational models inspired by biology have also been proposed for the time-delay estimation~\cite{sargolzaei2016sensorimotor}.
However, these models assume knowledge of the sensorimotor dynamics.  


\section{Discussion}
\subsection{Summary}
In this paper, we presented a self-organized hierarchical neural architecture for sensorimotor delay compensation in robots. 
In particular, we evaluated the proposed architecture in an imitation scenario, in which a simulated robot had to learn and reproduce visually demonstrated arm movements. 
Visuomotor sequences were extracted in the form of joint angles, which can be computed from a body skeletal representation in a straightforward way. 
Sequences generated by multiple users were learned using hierarchically-arranged GWR networks equipped with an increasingly large temporal window. 

The prediction of the visuomotor sequences was obtained by extending the GWR learning algorithm with a mapping mechanism of input and output vectors lying in the spatiotemporal domain. 
We conducted experiments with a dataset of 10 arm movement sequences showing that our system achieves low prediction error values on the training data and can adapt to unseen sequences in an online manner. 
Experiments also showed that a possible system malfunction causing loss of data samples has a relatively low impact on the overall performance of the system. 

\subsection{Growing self-organization and hierarchical learning}
The building block of our architecture is
the GWR network~\cite{marsland2002self}, which belongs to the unsupervised competitive learning class of artificial neural networks. 
A widely known algorithm of this class is the SOM~\cite{kohonen1993self}. 
The main component of these algorithms are the neurons equipped with weight vectors of a dimensionality equal to the input size. 
Through learning, the neurons become prototypes of the input space while preserving the input's topology, i.e., similar inputs are mapped to neurons that are near to each other. 
In the case of SOMs, these neurons are distributed and remain fixed in a 2D or a 3D lattice which has to be defined a priori and requires an optimal choice of its size.
In the GWR network, the topological structure of the neurons is dynamic and grows to adapt to the topological properties of the input space.
In this regard, the GWR network is similar to the GNG algorithm~\cite{fritzke1995growing}, another widely used growing self-organizing neural network.
However, the neural growth of the GWR  algorithm is not constant, as in the case of the GNG, but rather depends on how well the current state of the network represents the input data.
Thus, from the perspective of incremental learning, the GWR algorithm is more suitable than the GNG since new knowledge can be added to the network as soon as new data become available.

The hierarchical arrangement of the GWR networks equipped with a window in time memory is appealing due to the fact that it can dynamically change the topological structure in an unsupervised manner and learn increasingly more complex spatiotemporal dependencies of the data in the input space. 
This allows for a reuse of the neurons during sequence encoding, having learned prototypical spatiotemporal patterns out of the training sequences. 
Although this approach seems to be quite resource-efficient for the task of learning visuomotor sequences, the extent to which neurons are reused is tightly coupled with the input domain. 
In fact, in our experiments with input data samples represented as multi-dimensional vectors of both arms' shoulder and elbow angles, there was little to no overlap between training sequences. 
This led to a significant growth of the network with each sequence presentation.  

The parameters modulating the growth rate of each GWR network are the activation threshold and the firing counter threshold. 
The activation threshold $a_T$ establishes the maximum discrepancy between the input and the prototype neurons in the network. 
The larger we set the value of this parameter, the smaller is the discrepancy, i.e., the quantization error of the network. 
The firing counter threshold $f_T$ is used to ensure the training of recently added neurons before creating new ones.
Thus, smaller thresholds lead to more training of existing neurons and the slower creation of new ones, favoring again better network representations of the input.  
Intuitively, the less discrepancy between the input and the network representations, the smaller the inputs reconstruction error during prediction phase. 
However, less discrepancy means also more neurons. 
This proved to be not the main issue in our experiments since the number of neurons did not affect significantly the computation of the predicted values. 

A limitation of the sliding time-window technique for the encoding of temporal sequences is the high computational cost it introduces due to the data's higher dimensionality.
However, in our case using angles as body pose features leads to a low-dimensional input compared to e.g. images. 
So, the training with long time windows does not pose a computational challenge.
Furthermore, it has been shown that long-term predictions based on a sliding window are more accurate than recurrent approaches~\cite{2017arXiv170207486B}. 

The use of joint angles as visuomotor representations may seem to be a limitation of the proposed architecture due to the fact that it requires sensory input and robot actions to share the same representational space. 
For instance, in an object manipulation task, this requirement is not satisfied, since the visual feedback would be the position information given by the object tracking algorithm. 
This issue can be addressed by including both the position information and the corresponding robot joint angles as input to our architecture.
Due to the associative nature of self-organizing networks and their capability to function properly when receiving an incomplete input pattern, only the prediction of the object movement patterns would trigger the generation of corresponding patterns of the robot behavior.

\subsection{Future work}
An interesting direction for future work is the extension of the current implementation towards the autonomous generation of robot movements that account for both delay compensation as well as reaching a given action goal. 
For this purpose, the implementation of bidirectional Hebbian connections would have to be investigated in order to connect the last layer of the proposed architecture with a symbolic layer containing action labels~\cite{Parisi2016}\cite{schrodt2016just} and explore how such symbolic layer can modulate the generation of the movement patterns when diverging from the final goal. 

Future studies with the real robot will address the introduction of overall body configuration constraints for learning the perceived motion. 
The visual body tracking framework becomes unreliable in certain conditions, e.g., when the demonstrator is sitting or is touching objects in the background. 
In these cases, the provided body configurations may become unrealistic and cannot be mapped to the robot, or, even worse, when mapped to the robot may lead to a hardware break. 
For this reason, outlier detection mechanisms should be investigated in order to discard these unrealistic body configurations during training.

The imitation scenario studied in this paper was carried out offline, i.e., the synchronization was evaluated on an acquired data set of motion patterns.
However, the successful application of the proposed learning algorithm in providing accurate motor commands, thereby compensating the sensorimotor delay, encourages future experiments comprising an HRI user study in which participants will be able to teach the motion patterns directly to the robot.

\section*{Acknowledgment}
The authors gratefully acknowledge partial support by the EU- and City of Hamburg-funded program Pro-Exzellenzia 4.0, the German
Research Foundation DFG under project CML (TRR 169), and the Hamburg
Landesforschungsf\"orderungsprojekt.

\bibliographystyle{IEEEtran}
\bibliography{biblio}
\begin{IEEEbiography}
[{\includegraphics[width=1in,height=1.25in,clip,keepaspectratio]{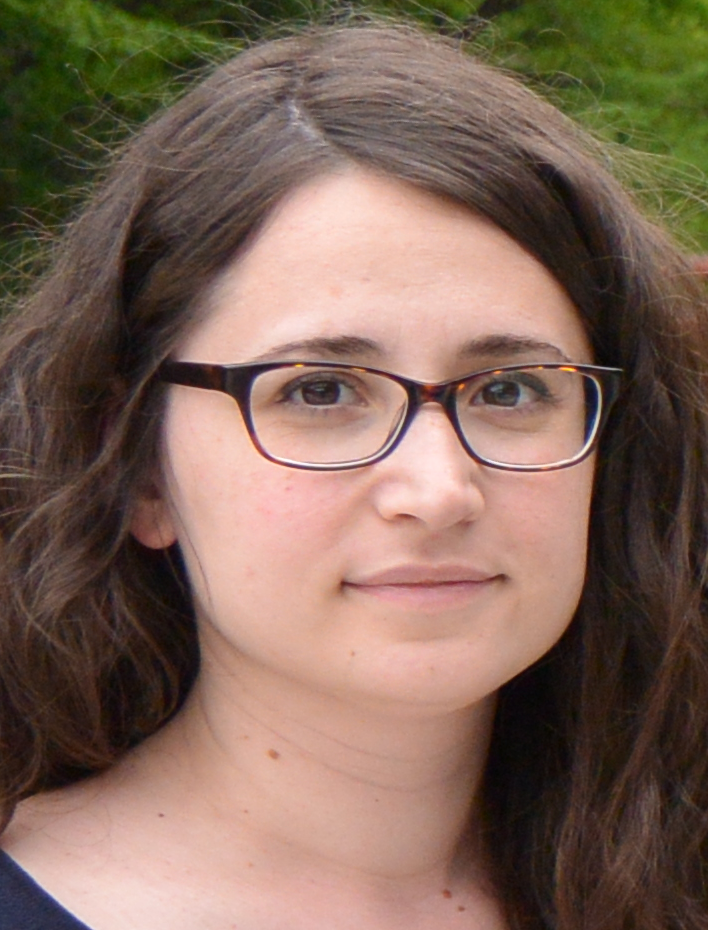}}]{Luiza Mici}
received her Bachelor's and Master's degree in Computer Engineering from the University of Siena, Italy. 
Since 2015, she is a research associate and PhD candidate in the Knowledge Technology Group at the University of Hamburg, Germany, where she was part of the research project CML (Crossmodal Learning).
Her main research interests include perception and learning, neural network self-organization, and bio-inspired action recognition.
\end{IEEEbiography}
\begin{IEEEbiography}
[{\includegraphics[width=1in,height=1.25in,clip,keepaspectratio]{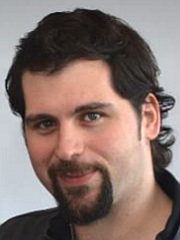}}]{German I. Parisi}
received his Bachelor’s and Master’s degree in Computer Science from the University of Milano-Bicocca, Italy.
In 2017 he received his PhD in Computer Science from the University of Hamburg, Germany, where he was part of the research project CASY (Cognitive Assistive Systems) and the international PhD research training group CINACS (Cross-Modal Interaction in Natural and Artificial Cognitive Systems).
In 2015 he was a visiting researcher at the Cognitive Neuro-Robotics Lab at the Korea Advanced Institute of Science and Technology (KAIST), Daejeon, South Korea.
Since 2016 he is a research associate of international project Transregio TRR 169 on Crossmodal Learning in the Knowledge Technology Institute at the University of Hamburg, Germany.
His main research interests include neurocognitive systems for human-robot assistance, computational models for multimodal integration, neural network self-organization, and deep learning.
\end{IEEEbiography}
\begin{IEEEbiography}[{\includegraphics[width=1in,height=1.25in,clip,keepaspectratio]{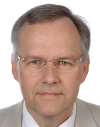}}]{Stefan Wermter}
is Full Professor at the University of Hamburg and Director of the Knowledge Technology institute. He holds an MSc from the University of Massachusetts in Computer Science, and a PhD and Habilitation in Computer Science from the University of Hamburg. He has been a research scientist at the International Computer Science Institute in Berkeley, US before leading the Chair in Intelligent Systems at the University of Sunderland, UK. His main research interests are in the fields of neural networks, hybrid systems, neuroscience-inspired computing, cognitive robotics and natural communication. He has been general chair for the International Conference on Artificial Neural Networks 2014. He is an associate editor of the journals “Transactions of Neural Networks and Learning Systems”, “Connection Science”, “International Journal for Hybrid Intelligent Systems” and “Knowledge and Information Systems” and he is on the editorial board of the journals “Cognitive Systems Research”, “Cognitive Computation” and “Journal of Computational Intelligence”. Currently he serves as Co-coordinator of the DFG-funded SFB/Transregio International Collaborative Research Centre on “Crossmodal Learning” and is coordinator of the European Training Network SECURE on Safe Robots.
\end{IEEEbiography}

\end{document}